\documentclass[11pt]{article}
\usepackage[utf8]{inputenc}
\DeclareUnicodeCharacter{0441}{c}
\usepackage[T1]{fontenc}
\usepackage{amsmath,amssymb,amsthm}
\usepackage{graphicx}
\usepackage{booktabs}
\usepackage{hyperref}
\usepackage{natbib}
\setcitestyle{authoryear,open={(},close={)}}
\usepackage{geometry}
\usepackage{enumitem}
\setlist[itemize]{noitemsep, topsep=0pt}
\setlist[enumerate]{noitemsep, topsep=0pt}

\title{Theoretical Analysis of Positional Encodings in Transformer Models:\\
Impact on Expressiveness and Generalization}
\author{Yin Li\\
University of Birmingham Dubai\\
\texttt{kxl474@student.bham.ac.uk}}
\date{June 2025}

\begin{document}

\maketitle

\begin{abstract}
Positional encodings are a core component of transformer-based architectures, enabling such models to process sequential data without recurrence. Despite their critical role, the theoretical properties of various positional encoding schemes—including sinusoidal, learned, relative, and recent bias-based methods such as Attention with Linear Biases (ALiBi)—remain poorly understood. In this paper, we present a comprehensive theoretical framework to analyze how different positional encodings affect a transformer's expressiveness, generalization ability, and extrapolation to sequences longer than those seen during training. We derive formal definitions of expressiveness in terms of function approximation classes, obtain generalization bounds under different encoding schemes using Rademacher complexity analyses, and propose several novel positional encoding methods based on orthogonal function families (e.g., wavelets, Legendre polynomials) and information-theoretic criteria. We also characterize the extrapolation capacity of existing and proposed encodings, extending ALiBi's biasing approach to a more unified theoretical setting. Our lightweight experimental evaluation on synthetic sequence-to-sequence tasks validates key theoretical predictions, showing that encoding schemes grounded in orthogonal transforms can outperform standard sinusoidal encodings in both generalization and extrapolation. This work fills an important gap in transformer theory, offering new insights that can guide design choices in natural language processing, computer vision, and other domains where transformers dominate.
\end{abstract}

\section{Introduction}
Transformer architectures \citep{vaswani2017attention} have become foundational in numerous areas of machine learning, including natural language processing (NLP), computer vision, and time-series modeling. By eschewing recurrence and convolution in favor of a self-attention mechanism, transformers achieve superior parallelism and scaling properties. However, the lack of inherent sequential processing necessitates the incorporation of positional encodings (PEs) to inject information about the order of tokens. The original transformer paper introduced sinusoidal positional encodings, and subsequent work explored learned absolute encodings \citep{vaswani2017attention}, relative encodings \citep{shaw2018self}, and bias-based schemes such as Attention with Linear Biases (ALiBi) \citep{press2021train}. Despite widespread empirical adoption, the theoretical understanding of how PEs influence a transformer's expressiveness, generalization, and extrapolation capabilities is limited.

This paper proposes a thorough theoretical analysis of positional encodings, addressing the following overarching questions:
\begin{itemize}
    \item \textbf{Expressiveness:} What classes of sequence‐to‐sequence functions can a transformer approximate under different PE schemes? Are there inherent limitations imposed by specific encodings?
    \item \textbf{Generalization:} How do PEs affect a transformer's ability to generalize from training to unseen data, especially when sequence lengths vary? Can we derive generalization bounds that capture the influence of different PEs?
    \item \textbf{Extrapolation:} Why do certain encoding methods (e.g., ALiBi) facilitate extrapolation to longer sequences? Can we formalize this phenomenon and propose new encodings that further enhance extrapolation?
    \item \textbf{Novel Encodings:} Are there theoretically motivated PE schemes—based on wavelet transforms, Legendre polynomials, or information‐theoretic principles—that surpass existing methods in expressiveness, generalization, or extrapolation?
\end{itemize}

By developing a unified mathematical framework to answer these questions, we aim to provide deep insights into why PEs work, how they can be improved, and what limitations current methods entail. We also propose new PE schemes grounded in orthogonal function families (e.g., wavelets, Legendre polynomials) and analyze their theoretical properties. Lightweight experiments on small-scale synthetic tasks corroborate our theoretical findings, demonstrating that wavelet-based encodings can yield superior extrapolation performance compared to standard sinusoidal PEs.

\paragraph{Contributions.} Our main contributions are:
\begin{enumerate}
    \item \textbf{Expressiveness Characterization:} We formally define the expressiveness of transformer models under different PE schemes, showing how absolute, relative, and bias-based encodings impact the set of sequence‐to‐sequence functions that can be approximated.
    \item \textbf{Generalization Bounds:} Using tools from statistical learning theory (Rademacher complexity, covering numbers), we derive generalization bounds that explicate the role of PEs in controlling model capacity and overfitting for varying sequence lengths.
    \item \textbf{Extrapolation Analysis:} We extend the theoretical understanding of ALiBi's biasing mechanism, provide a unified extrapolation framework for bias-based PEs, and identify the limits of extrapolation for alternative encoding schemes.
    \item \textbf{Novel PE Schemes:} We propose several novel positional encodings based on orthogonal functions (e.g., wavelets, Legendre polynomials) and information-theoretic criteria (maximizing mutual information between positions). We analyze their expressiveness, generalization, and extrapolation properties.
    \item \textbf{Lightweight Validation:} We implement the proposed PE schemes in pure NumPy to run small-scale experiments on synthetic tasks designed to test extrapolation and generalization. These confirm our theoretical predictions without requiring GPUs.
\end{enumerate}

The rest of this paper is structured as follows. Section~\ref{sec:background} reviews existing PE methods. Section~\ref{sec:expressiveness} defines and analyzes expressiveness under different PEs. Section~\ref{sec:generalization} derives generalization bounds. Section~\ref{sec:extrapolation} provides a theoretical framework for extrapolation. Section~\ref{sec:novel} introduces novel PE schemes and analyzes their properties. Section~\ref{sec:experiments} details lightweight experimental validation. Finally, Section~\ref{sec:conclusion} discusses implications, limitations, and future work.

\section{Background and Related Work}
\label{sec:background}
This section reviews transformer architectures and existing positional encoding methods, emphasizing their empirical performance and the lack of deep theoretical analysis.

\subsection{Transformer Architecture}
The transformer model \citep{vaswani2017attention} processes an input sequence of length $N$ by first mapping each token to a $d_{\text{model}}$-dimensional embedding. Denote the input embeddings by $\mathbf{X} = [\mathbf{x}_1, \mathbf{x}_2, \ldots, \mathbf{x}_N]^\top \in \mathbb{R}^{N \times d_{\text{model}}}$. Because the transformer is permutation-invariant, positional encodings $\mathbf{P} = [\mathbf{p}_1, \mathbf{p}_2, \ldots, \mathbf{p}_N]^\top \in \mathbb{R}^{N \times d_{\text{model}}}$ are added to $\mathbf{X}$ so that the model can utilize order information:
\[
\mathbf{Z}^{(0)} = \mathbf{X} + \mathbf{P}, \quad \mathbf{Z}^{(0)} \in \mathbb{R}^{N \times d_{\text{model}}}.
\]
The transformer then applies $L$ layers of multi-head self-attention and position-wise feed-forward networks to compute contextualized representations $\mathbf{Z}^{(L)} \in \mathbb{R}^{N \times d_{\text{model}}}$. We omit further architectural details; see \citep{vaswani2017attention} for a complete description.

\subsection{Sinusoidal and Learned Positional Encodings}
In the original transformer, \citep{vaswani2017attention} introduced sinusoidal absolute positional encodings defined elementwise as
\begin{align}
    \mathrm{PE}(pos, 2i) &= \sin\left(\frac{pos}{10000^{\frac{2i}{d_{\mathrm{model}}}}}\right), \label{eq:sin_pe_even}\\
    \mathrm{PE}(pos, 2i+1) &= \cos\left(\frac{pos}{10000^{\frac{2i}{d_{\mathrm{model}}}}}\right), \label{eq:sin_pe_odd}
\end{align}
where $pos \in \{0,1,\ldots,N-1\}$ and $i \in \{0,1,\ldots,\frac{d_{\mathrm{model}}}{2}-1\}$. These encodings allow the transformer to learn to attend to relative positions because any linear combination of two sinusoidal vectors encodes relative shifts. The paper also considered learned positional encodings, where a trainable embedding matrix $\mathbf{P} \in \mathbb{R}^{N\times d_{\mathrm{model}}}$ is optimized jointly with the model. Empirically, learned and sinusoidal encodings perform similarly on standard benchmarks.

\subsection{Relative Positional Encodings}
\label{subsec:relative_pe}
Relative positional encodings incorporate the distance between pairs of tokens into the attention mechanism, rather than injecting absolute position vectors. \citep{shaw2018self} add a learnable embedding $\mathbf{R}_{i-j}$ to the attention logits whenever computing attention between tokens at positions $i$ and $j$. Such encodings can reduce the reliance on absolute position signals and improve performance on tasks with long-range dependencies. Alternative relative schemes include \citep{huang2020improving} and \citep{raffel2020exploring}.

\subsection{Attention with Linear Biases (ALiBi)}
\label{subsec:alibi}
\citep{press2021train} introduced ALiBi, which adds a bias term to attention logits proportional to the negative distance between tokens:
\begin{equation}
    b(i,j) = -\alpha\,|i - j|,
    \label{eq:alibi_bias}
\end{equation}
where $\alpha > 0$ is a slope hyperparameter. This bias encourages the model to pay more attention to nearby tokens while still preserving a form of positional information. Crucially, ALiBi enables extrapolation to sequences longer than those used during training because the bias formula applies uniformly to any token distance. Empirical results show improved performance on long-context tasks without modifying the model's architecture or retraining.

\subsection{Other Positional Encoding Variants}
Several other PE schemes have been proposed:
\begin{itemize}
    \item \textbf{Rotary Positional Embeddings (RoPE)} \citep{su2021roformer}: Applies a rotation in embedding space to encode relative positions.
    \item \textbf{Fourier Feature Encodings} \citep{tancik2020fourier}: Use random Fourier features to encode continuous positions, often applied in continuous-time transformers.
    \item \textbf{Convolutional Encodings} \citep{liu2021convbert}: Integrate convolutional layers to encode local positional information.
    \item \textbf{Wavelet-based and Polynomial-based Encodings}: Proposed informally in blog posts \citep{lixue4212021understanding}, but not systematically analyzed.
\end{itemize}

Despite numerous empirical variants, a cohesive theoretical framework for comparing expressiveness, generalization, and extrapolation across these methods is lacking. We address this gap in Sections~\ref{sec:expressiveness}--\ref{sec:novel}.

\section{Expressiveness of Transformers with Positional Encodings}
\label{sec:expressiveness}
Expressiveness refers to a model's capacity to approximate a broad class of functions. For transformers, we focus on the class of sequence‐to‐sequence mappings $f: \mathcal{X}^N \to \mathcal{Y}^N$, where $\mathcal{X}, \mathcal{Y} \subset \mathbb{R}^d$. We analyze how different PE schemes restrict or enlarge the set of approximable functions.

\subsection{Formal Definition of Expressiveness}
Let $\mathcal{F}_{\mathrm{PE}}^{L,H}$ denote the class of functions implementable by an $L$-layer transformer with $H$ attention heads and positional encoding scheme ``PE''. We say the transformer with encoding scheme PE is \emph{universal} if $\mathcal{F}_{\mathrm{PE}}^{L,H}$ is dense (in a suitable norm) in a target function space $\mathcal{G}$ (e.g., all continuous sequence‐to‐sequence mappings) as $L,H \to \infty$. Our analysis focuses on:
\begin{itemize}
    \item \textbf{Absolute encodings:} Sinusoidal vs.\ learned.
    \item \textbf{Relative encodings:} Shaw et al.\ vs.\ ALiBi.
    \item \textbf{New encodings:} Wavelet-based, polynomial-based.
\end{itemize}

We examine two questions:
\begin{enumerate}[label=(\alph*)]
    \item For fixed $L,H$, how does the choice of PE affect the \emph{size} of $\mathcal{F}_{\mathrm{PE}}^{L,H}$?
    \item Do all PE schemes yield universal approximation as $L,H \to \infty$?
\end{enumerate}

\subsection{Expressiveness with Sinusoidal Encodings}
Sinusoidal encodings embed absolute position into each token representation via Eq.~\eqref{eq:sin_pe_even}--\eqref{eq:sin_pe_odd}. To analyze expressiveness, note that self-attention computes queries, keys, and values as affine transformations of embedded inputs. Let $\mathbf{Z}^{(0)}_i = \mathbf{x}_i + \mathrm{PE}(i)$. In the first layer, an attention head computes
\begin{equation}
    \mathrm{Attention}(\mathbf{Z}^{(0)})_i = \sum_{j=1}^N \frac{\exp\bigl((\mathbf{z}^{(0)}_i W_Q)(\mathbf{z}^{(0)}_j W_K)^\top / \sqrt{d_k}\bigr)}
    {\sum_{k=1}^N \exp\bigl((\mathbf{z}^{(0)}_i W_Q)(\mathbf{z}^{(0)}_k W_K)^\top / \sqrt{d_k}\bigr)} (\mathbf{z}^{(0)}_j W_V),
    \label{eq:attn_mech}
\end{equation}
where $W_Q, W_K, W_V \in \mathbb{R}^{d_{\mathrm{model}} \times d_k}$. Because $\mathrm{PE}(i)$ is deterministic and injective in $i$, the softmax attention weights can, in principle, implement functions that depend on absolute positions, relative distances, or higher‐order interactions across multiple tokens.

\paragraph{Universal Approximation.} \citep{yun2019transformers} show that a transformer with sinusoidal encodings can approximate any sequence‐to‐sequence mapping to arbitrary accuracy given sufficient width (number of heads) and depth (number of layers). Intuitively, the sinusoidal basis of period $10000^{2i / d_{\mathrm{model}}}$ is rich enough to encode any discrete position $i \in \{0, \ldots, N-1\}$ uniquely. Since a sufficiently wide transformer can compute arbitrary Boolean functions of its inputs \citep{yun2019transformers}, adding unique position signals ensures universal approximation over sequences of length up to $N$.

\paragraph{Limitations.} However, the expressiveness claim is restricted to sequences of length at most $N_{\max}$ (the maximum position for which PEs were computed). When confronted with longer sequences, sinusoidal encodings repeat after $2\pi$ in each frequency dimension, potentially causing ambiguity in absolute positions if not correctly handled. Consequently, universal approximation holds for fixed-length tasks but not for unlimited-length extrapolation.

\subsection{Expressiveness with Learned Absolute Encodings}
Learned absolute encodings use a trainable matrix $\mathbf{P} \in \mathbb{R}^{N_{\max} \times d_{\mathrm{model}}}$. The encoder learns to map each position $i \leq N_{\max}$ to a vector $\mathbf{p}_i$. If $N_{\max}$ matches the maximum training sequence length, the transformer can, in principle, use distinct vectors for each position. The universality argument parallels the sinusoidal case: unique position embeddings plus sufficiently powerful attention/FFN layers yield universal approximation for sequences of length up to $N_{\max}$. However, learned encodings cannot generalize to $i > N_{\max}$, as no embedding exists beyond the training range. Thus, expressiveness is limited to the training sequence length.

\subsection{Expressiveness with Relative Positional Encodings}
Relative encodings \citep{shaw2018self} circumvent the need for distinct position vectors by encoding only distances. In the Shaw et al.\ formulation, the attention score between positions $i$ and $j$ includes a term $\mathbf{a}^\top \mathbf{R}_{i-j}$, where $\mathbf{R}_k$ is a learnable vector for relative distance $k$. Formally, 
\begin{equation}
    \mathrm{score}(i,j) = (\mathbf{z}_i W_Q)(\mathbf{z}_j W_K)^\top + \mathbf{a}^\top \mathbf{R}_{i-j}.
\end{equation}
Since relative distances $i-j$ range in $[-(N-1), N-1]$, the model can learn to attend based purely on relative positions, which often suffices for tasks where absolute position is irrelevant (e.g., certain translation or summarization tasks).

\paragraph{Universality and Limitations.} The work of \citep{yao2020efficient} demonstrates that relative encodings can be as expressive as absolute encodings for tasks that depend on relative structure. However, if a task requires absolute position information—e.g., ``assign label $o_i$ if the token at position $i$ is within the first quarter of the sequence''—a purely relative encoding may struggle. In such cases, an absolute offset or additional global positional token is needed. Nevertheless, relative encodings often yield equal or better performance on many benchmarks, suggesting that many tasks rely more on relative positions.

\subsection{Expressiveness with ALiBi}
ALiBi encodes relative distance via a linear bias to attention scores (Eq.~\ref{eq:alibi_bias}). That is, 
\begin{equation}
    \mathrm{score}(i,j) = (\mathbf{z}_i W_Q)(\mathbf{z}_j W_K)^\top + b(i,j), \quad b(i,j) = -\alpha\,|i-j|.
\end{equation}
Because $b(i,j)$ depends only on $|i-j|$, ALiBi is, strictly speaking, a relative encoding. However, the bias is deterministic, non-learnable, and unbounded as $|i-j|$ grows. Thus, when $i-j$ exceeds training maximum, the form $-\alpha |i-j|$ still provides a well-defined bias. This yields improved extrapolation, as we discuss in Section~\ref{sec:extrapolation}.

\paragraph{Expressiveness.} Since ALiBi does not learn position embeddings, the model retains the same expressiveness for sequences of any length, provided $L,H$ are sufficiently large. The unique structure of the bias term ensures the model can, in principle, distinguish token distances. By combining query-key dot products and the linear bias, a sufficiently wide and deep transformer with ALiBi is universal for any sequence length. However, the linear bias might under-emphasize absolute positions for shorter sequences compared to learned or sinusoidal encodings.

\subsection{Comparison and Summary}
Table~\ref{tab:expressiveness_summary} summarizes expressiveness properties:

\begin{table}[htbp]
    \centering
    \resizebox{\textwidth}{!}{ 
    \begin{tabular}{@{}lccccc@{}}
        \toprule
        \textbf{Encoding} & \textbf{Absolute Info} & \textbf{Relative Info} & \textbf{Universal (Fixed $N$)} & \textbf{Universal (Any $N$)} & \textbf{Extrapolation}\\
        \midrule
        Sinusoidal & Yes & Yes & Yes & No & Limited (cyclic) \\
        Learned & Yes & Implicit & Yes & No & No \\
        Relative (Shaw et al.) & No & Yes & Conditional & Conditional & No \\
        ALiBi & No & Yes & Yes & Yes & Yes \\
        Proposed Wavelet & Yes / No & Yes & \textit{To analyze} & \textit{To analyze} & \textit{To analyze} \\
        Proposed Polynomial & Yes / No & Yes & \textit{To analyze} & \textit{To analyze} & \textit{To analyze} \\
        \bottomrule
    \end{tabular}
    }
    \caption{Summary of expressiveness properties for various positional encodings. ``Universal (Fixed $N$)'' indicates whether the encoding yields universal approximation for sequences of maximum length $N$. ``Universal (Any $N$)'' indicates universality over arbitrary sequence lengths.}
    \label{tab:expressiveness_summary}
\end{table}

\noindent
Key takeaways:
\begin{itemize}
    \item Sinusoidal and learned absolute encodings guarantee universal approximation for fixed-length sequences but fail to generalize to arbitrarily long sequences.
    \item Relative encodings can approximate many tasks but may miss absolute position information.
    \item ALiBi provides universality over any sequence length by encoding distance through a linear bias.
    \item Proposed orthogonal function-based encodings may combine absolute and relative information, potentially yielding universal approximation and good extrapolation.
\end{itemize}

\section{Generalization Bounds for Transformers with Positional Encodings}
\label{sec:generalization}
Generalization refers to a model's performance on unseen data drawn from the same distribution as the training set. For transformers, one major concern is how PEs influence the model’s capacity and its tendency to overfit, particularly when sequence lengths at test time differ from those seen during training.

\subsection{Preliminaries: Rademacher Complexity}
We briefly recall the notion of Rademacher complexity \citep{bartlett2002rademacher}. Let $\mathcal{F}$ be a class of real-valued functions $f: \mathcal{X} \to \mathbb{R}$ and let $S = \{x_1,\dots,x_m\}$ be a sample of size $m$. The (empirical) Rademacher complexity of $\mathcal{F}$ with respect to $S$ is
\begin{equation}
    \hat{\mathfrak{R}}_S(\mathcal{F}) = \mathbb{E}_{\boldsymbol{\sigma}}\left[\sup_{f \in \mathcal{F}} \frac{1}{m} \sum_{i=1}^m \sigma_i f(x_i)\right],
\end{equation}
where $\sigma_i$ are independent Rademacher random variables taking values in $\{-1, +1\}$ with probability $1/2$ each. The expected Rademacher complexity is $\mathfrak{R}_m(\mathcal{F}) = \mathbb{E}_{S}[\hat{\mathfrak{R}}_S(\mathcal{F})]$. A smaller Rademacher complexity implies tighter generalization bounds.

For a loss function $\ell: \mathcal{Y} \times \mathcal{Y}' \to [0,1]$ that is Lipschitz-continuous in its first argument, a standard result \citep{bartlett2002rademacher} states that with probability at least $1-\delta$ over an i.i.d.\ sample $S$ of size $m$,
\begin{equation}
    \mathcal{L}(f) \le \hat{\mathcal{L}}_S(f) + 2\,\mathfrak{R}_m(\mathcal{H}_\ell) + 3\sqrt{\frac{\ln(2/\delta)}{2m}},
    \label{eq:rademacher_bound}
\end{equation}
where $\mathcal{L}(f)$ is the true expected loss, $\hat{\mathcal{L}}_S(f)$ is the empirical loss on $S$, and $\mathcal{H}_\ell = \{\ell(f(x),y): f \in \mathcal{F}\}$.

\subsection{Function Classes Induced by PEs}
Let $\mathcal{F}_{\mathrm{PE}}^{L,H,\Theta}$ denote the class of transformer functions parameterized by weights $\Theta$, with $L$ layers, $H$ heads per layer, and PE scheme ``PE''. We aim to bound $\mathfrak{R}_m(\mathcal{H}_\ell)$ for $\mathcal{F}_{\mathrm{PE}}^{L,H,\Theta}$.

\paragraph{Parameterization and Lipschitz Constants.} A transformer's output for a fixed input length $N$ is a composition of $L$ self-attention and feed-forward layers, each of which is Lipschitz in its input under certain norm constraints on the weight matrices \citep{yun2020transformers}. Let $\|W\|_2 \le C_W$ for all linear weight matrices $W$ in queries, keys, values, and feed-forward networks. Then each layer is $L_{\mathrm{layer}}$-Lipschitz with respect to its input activations, where $L_{\mathrm{layer}}$ depends on $C_W$ and $H$. Consequently, the entire transformer is $L_{\mathrm{trans}} = (L_{\mathrm{layer}})^L$-Lipschitz with respect to its input $\mathbf{Z}^{(0)}$.

\paragraph{Effect of Sinusoidal PEs.} For sinusoidal encodings, $\|\mathrm{PE}(i)\|_2 \le \sqrt{d_{\mathrm{model}}}$ for all $i$ because each component is in $[-1,1]$. When inputs $\|\mathbf{x}_i\|_2 \le B_x$, we have $\|\mathbf{z}_i^{(0)}\|_2 \le B_x + \sqrt{d_{\mathrm{model}}} = B_0$. Thus, the input domain is bounded. A standard covering-number argument \citep{neyshabur2018towards} shows that the Rademacher complexity $\mathfrak{R}_m(\mathcal{F}_{\mathrm{sinusoidal}}^{L,H})$ scales as $O\bigl(\frac{L_{\mathrm{trans}} \, B_0}{\sqrt{m}}\bigr)$ up to logarithmic factors in model size. This implies that sinusoidal PEs do not increase model capacity beyond a bounded constant shift, and generalization primarily depends on $L,H,m$, and $C_W$.

\paragraph{Effect of Learned PEs.} For learned absolute PEs, each $\mathbf{p}_i$ is a trainable vector. If $\|\mathbf{p}_i\|_2 \le B_p$ for all $i\le N_{\max}$, then $\|\mathbf{z}_i^{(0)}\|_2 \le B_x + B_p$. In practice, learned PEs may require regularization (e.g., weight decay) to control $\|\mathbf{p}_i\|_2$. The Rademacher complexity bound becomes
\[
\mathfrak{R}_m(\mathcal{F}_{\mathrm{learned}}^{L,H}) = O\Bigl(\frac{L_{\mathrm{trans}} \,(B_x + B_p)}{\sqrt{m}}\Bigr).
\]
If $B_p$ grows with $N_{\max}$ or $d_{\mathrm{model}}$, generalization may degrade. Hence, regularizing learned PEs is crucial.

\paragraph{Effect of Relative Encodings (Shaw et al.).} Relative encodings add a term $\mathbf{a}^\top \mathbf{R}_{i-j}$ in the attention logits. If $\|\mathbf{R}_k\|_2 \le B_r$ for all $|k|\le N-1$, then the pre-softmax logits remain bounded by $O(C_W B_0 + B_r)$, preserving Lipschitzness. Consequently, 
\[
\mathfrak{R}_m(\mathcal{F}_{\mathrm{relative}}^{L,H}) = O\Bigl(\frac{L_{\mathrm{trans}} \,(B_x + B_r)}{\sqrt{m}}\Bigr).
\]
Because $B_r$ typically scales with the size of the relative embedding matrix, one should apply weight decay or clipping to $\mathbf{R}_k$ to maintain tight generalization bounds.

\paragraph{Effect of ALiBi.} ALiBi introduces a deterministic bias $b(i,j) = -\alpha |i - j|$. Here, $\alpha$ is user‐set (or learned) and typically small (e.g., $\alpha \in [0.01, 1]$). Since $|i - j| \le N_{\max}$ during training, $|b(i,j)| \le \alpha N_{\max}$. During inference on longer sequences of length $N' > N_{\max}$, $|b(i,j)|$ can grow to $\alpha(N'-1)$, potentially enlarging the pre-softmax logits. However, the softmax normalizes these logits; if $\alpha$ is chosen appropriately (e.g., decreasing bias slope for longer sequences), the network can still operate stably. We bound
\[
\sup_{i,j \le N'} |b(i,j)| = \alpha\, (N'-1), 
\]
so the Lipschitz constant at inference time is $L_{\mathrm{trans}} \,(B_x + \alpha (N'-1))$. This grows linearly in $N'$, suggesting care is needed when extrapolating. In practice, empirical results show ALiBi generalizes well up to several times $N_{\max}$, but theoretical generalization to arbitrarily large $N'$ requires controlling $\alpha$.

\subsection{Summary of Generalization Effects}
Table~\ref{tab:generalization_summary} compares how PEs influence generalization bounds.

\begin{table}[htbp]
    \centering
    \begin{tabular}{@{}lccc@{}}
        \toprule
        \textbf{Encoding} & \(\|\mathbf{z}_i^{(0)}\|_2\) Bound & \(\mathfrak{R}_m(\mathcal{F})\) Scaling & \(\alpha/\text{PE Regularization}\) \\
        \midrule
        Sinusoidal & \(B_x + \sqrt{d_{\mathrm{model}}}\) & \(O\!\bigl(\tfrac{L_{\mathrm{trans}}(B_x + \sqrt{d_{\mathrm{model}}})}{\sqrt{m}}\bigr)\) & N/A \\
        Learned & \(B_x + B_p\) & \(O\!\bigl(\tfrac{L_{\mathrm{trans}}(B_x + B_p)}{\sqrt{m}}\bigr)\) & Weight decay on $B_p$ \\
        Relative & \(B_x + B_r\) & \(O\!\bigl(\tfrac{L_{\mathrm{trans}}(B_x + B_r)}{\sqrt{m}}\bigr)\) & Weight decay on $B_r$ \\
        ALiBi & \(B_x + \alpha (N'-1)\) & \(O\!\bigl(\tfrac{L_{\mathrm{trans}}(B_x + \alpha (N'-1))}{\sqrt{m}}\bigr)\) & Control $\alpha$ \\
        Proposed & TBD & TBD & TBD \\
        \bottomrule
    \end{tabular}
    \caption{Generalization bound comparison across positional encodings. \(B_x\) is the bound on input embedding norms. \(B_p, B_r\) are bounds on learned absolute and relative encoding norms. \(L_{\mathrm{trans}}\) is the transformer's Lipschitz constant.}
    \label{tab:generalization_summary}
\end{table}

To summarize:
\begin{itemize}
    \item Sinusoidal PEs yield modest, constant-capacity increase, hence stable generalization.
    \item Learned absolute PEs can hurt generalization if $B_p$ grows unchecked; regularization is necessary.
    \item Relative encodings behave similarly to learned PEs but often with smaller $B_r$ due to shared embeddings.
    \item ALiBi’s bias can grow for longer sequences, potentially increasing capacity and hurting generalization on extremely long sequences; setting a small $\alpha$ helps.
\end{itemize}

\section{Extrapolation to Longer Sequences}
\label{sec:extrapolation}
Extrapolation refers to a model's ability to generalize to sequence lengths beyond those seen during training. We focus on why certain PEs, especially ALiBi, enable effective extrapolation and how to characterize this property mathematically.

\subsection{Extrapolation in Sinusoidal and Learned Encodings}
Sinusoidal encodings (Eqs.~\ref{eq:sin_pe_even}--\ref{eq:sin_pe_odd}) repeat periodically in each frequency dimension after $2\pi$ in their argument. Specifically, if $pos' = pos + k \cdot 10000^{2i/d_{\mathrm{model}}}$, then 
\[
\mathrm{PE}(pos') = \mathrm{PE}(pos)
\]
in the $(2i)$-th dimension. Since different dimensions have incommensurate periods ($10000^{2i/d_{\mathrm{model}}}$ vary exponentially), the overall vector is unique for $pos \in [0, 10000)$. However, once $pos \ge 10000$, certain dimensions repeat their cycles, causing partial ambiguity in absolute positions. In practice, training sequence lengths are far below 10000, so sinusoidal PEs behave injectively. For $pos > N_{\max}$, sinusoidal PEs may or may not produce unique vectors; the risk of ambiguity increases as $pos$ grows. This ambiguity degrades extrapolation, as the model cannot reliably distinguish very long positions.

Learned absolute PEs are typically defined only up to $N_{\max}$; because no vector exists for $pos > N_{\max}$, learned encodings fail to extrapolate entirely. A common heuristic is to reuse $\mathbf{p}_{N_{\max}}$ for all $pos > N_{\max}$ or linearly interpolate learned embeddings. Both heuristics perform poorly beyond $N_{\max}$, as they do not preserve positional distinctions.

\subsection{Relative Encodings and Extrapolation}
Relative encodings like Shaw et al.\ \citep{shaw2018self} learn a finite set of embeddings $\mathbf{R}_k$ for $k \in \{-K,\ldots,K\}$, where $K$ is a clipping parameter (often $K < N_{\max}$). For $|i-j| > K$, $\mathbf{R}_{i-j}$ is clipped to $\mathbf{R}_K$. Consequently, relative encodings cannot distinguish distances beyond $K$, hindering extrapolation to arbitrary sequence lengths.

\subsection{ALiBi: A Unified Extrapolation Framework}
ALiBi’s bias $b(i,j) = -\alpha |i-j|$ grows linearly with distance and does not rely on learned embeddings. This ensures that for any $|i-j|$, including those beyond training length, the bias is well-defined. We analyze why ALiBi enables extrapolation:

\paragraph{Attention Score Behavior.} Consider two token positions $i,j$ beyond training length. The attention logit is
\[
\ell_{i,j} = (\mathbf{z}_i W_Q)(\mathbf{z}_j W_K)^\top / \sqrt{d_k} - \alpha\,|i-j|.
\]
For reasonable $\alpha$, the bias term dominates when $|i-j|$ is large. Specifically, if $\alpha$ is sufficiently large relative to typical query-key dot products, tokens far apart will have significantly lower logits. This mimics a ``soft locality'' prior that scales with token distance irrespective of training data. Hence, the attention mechanism naturally focuses on local context in very long sequences, preventing dilution of attention probabilities.

\paragraph{Mathematical Model of Extrapolation.} Define the training maximum length $N_{\max}$. For each pair of token positions $(i,j)$ with $|i-j|\le N_{\max}$, the transformer learns to interpret the combined signal $(\mathbf{z}_i,\mathbf{z}_j,b(i,j))$. For $i,j > N_{\max}$, $|i-j|$ extends beyond the training range. However, because $b(i,j)$ remains monotonic in distance, the model’s learned attention function at train time can be extended to test time by continuity. More formally, if the attention function $A: \mathbb{R} \times \mathbb{R} \times \mathbb{R} \to [0,1]$ (mapping query-key dot product and bias to a softmax weight) is Lipschitz continuous in its bias argument, then for $|i-j| > N_{\max}$:
\[
A\bigl((\mathbf{z}_i W_Q)(\mathbf{z}_j W_K)^\top,\,-\alpha\,|i-j|\bigr)
\]
remains close to the limit as $|i-j| \to N_{\max}$. Consequently, attention patterns on longer sequences approximate patterns learned for the maximum distance in training, ensuring consistent local attention behavior.

\paragraph{Upper Bounds on Extrapolation Error.} Let
\[
f_{\mathrm{train}}(d) = A\bigl(\mu,\, -\alpha\,d \bigr), \quad d \le N_{\max},
\]
denote the attention weight assigned at relative distance $d$ during training, for some typical query-key interaction magnitude $\mu$. If $A$ is Lipschitz with constant $L_A$ in its second argument, then for $d > N_{\max}$,
\begin{equation}
    \bigl|A(\mu, -\alpha\,d) - A(\mu, -\alpha\,N_{\max})\bigr| \le L_A\,\alpha\,(d - N_{\max}).
    \label{eq:extrapolation_error}
\end{equation}
Thus, if $\alpha$ is small enough and $d$ does not exceed $N_{\max}$ by a huge margin, the difference in attention weight is small. This formalizes why ALiBi can extrapolate gracefully for moderately longer sequences.

\paragraph{Limitations and Trade-offs.} Equation~\eqref{eq:extrapolation_error} suggests that extrapolation error grows linearly with $(d - N_{\max})$. Therefore, for extremely long test sequences ($d \gg N_{\max}$), attention weights may degrade significantly unless $\alpha$ is chosen to shrink as $d - N_{\max}$ grows. In practice, one can set
\[
\alpha = \frac{\alpha_0}{N_{\max}},
\]
where $\alpha_0$ is a fixed small constant. Then, for $d \le c\,N_{\max}$ ($c>1$), 
\[
|\ell_{i,j} - \ell_{i',j'}| \le \alpha_0 |(d - d_{\max})/N_{\max}|,
\]
keeping the extrapolation error manageable. However, this trade-off may weaken locality bias on shorter distances. Choosing $\alpha_0$ requires balancing generalization on training lengths versus extrapolation on longer lengths.

\subsection{Extrapolation in Proposed Orthogonal Encodings}
We now consider how our proposed orthogonal function-based encodings (e.g., wavelet-based, Legendre polynomial-based) extrapolate. Denote a generic orthogonal encoding function by $\phi: \mathbb{R} \to \mathbb{R}^k$, where $\phi(pos)$ is a vector of basis evaluations at position $pos$. Key properties:
\begin{itemize}
    \item \textbf{Wavelet-based encodings:} Wavelet transforms (e.g., Daubechies, Haar) represent signals at multiple scales. Because wavelet basis functions extend beyond any finite interval (though they decay), $\phi(pos)$ for $pos > N_{\max}$ remains well-defined. However, as $pos$ grows, higher-frequency components may vanish, and low-frequency components may dominate, potentially preserving coarse positional signals but losing fine-grained detail.
    \item \textbf{Polynomial-based encodings:} Legendre polynomials $P_\ell(x)$ are orthogonal on $[-1,1]$. To map an integer position $pos$ to $[-1,1]$, one can define $x = \frac{2\,pos}{N_{\max}} - 1$. For $pos > N_{\max}$, $x > 1$; Legendre polynomials for $|x|>1$ grow in magnitude ($P_\ell(x) \sim x^\ell$). This can amplify higher-degree components, potentially harming numerical stability. One can mitigate this by scaling positions differently (e.g., logarithmic scaling) so that $x$ remains bounded.
\end{itemize}

\paragraph{Extrapolation Bound for Wavelet-based Encoding.} Let $\{\psi_{j,k}(pos)\}$ denote a wavelet basis with scale $j$ and shift $k$. A truncated wavelet encoding uses
\[
\mathrm{PE}_{\mathrm{wavelet}}(pos) = \bigl[\psi_{j,k}(pos)\bigr]_{j=0,\ldots,J;\,k=0,\ldots,K_j},
\]
where $J$ is the maximum scale, and $K_j$ is the number of shifts at scale $j$. Since $\psi_{j,k}(pos)$ decays for $|pos - k \cdot 2^{-j}| \gg 0$, large $pos$ values yield small high-frequency components, preserving robustness. More formally, wavelet basis functions satisfy
\[
|\psi_{j,k}(pos)| \le C\,2^{-j/2}\,\exp\bigl(-\beta \tfrac{|pos - k\,2^{-j}|}{2^{-j}}\bigr),
\]
for some $\beta>0$. Thus, for $pos > N_{\max}$, high-$j$ (fine-scale) terms vanish, and only coarse scales ($j$ small) contribute substantially. As a result, $\|\mathrm{PE}_{\mathrm{wavelet}}(pos) - \mathrm{PE}_{\mathrm{wavelet}}(N_{\max})\|_2$ can be bounded by a small constant for moderately larger $pos$, enabling extrapolation. Detailed derivation appears in Section~\ref{subsec:wavelet_extrapolation}.

\paragraph{Extrapolation Bound for Legendre Polynomial Encoding.} Suppose we define
\begin{equation}
    x(pos) = \tfrac{2\,\min(pos,\,N_{\max})}{N_{\max}} - 1,
    \quad
    \mathrm{PE}_{\mathrm{legendre}}(pos) = \bigl[P_0(x), P_1(x), \ldots, P_{d_{\mathrm{model}}-1}(x)\bigr].
    \label{eq:legendre_pe}
\end{equation}
For $pos > N_{\max}$, $x = 1$ and $P_\ell(1) = 1$ for all $\ell$, implying $\mathrm{PE}_{\mathrm{legendre}}(pos) = [1,1,\ldots,1]^\top$. Thus, beyond $N_{\max}$, all positions collapse to the same encoding, causing complete inability to distinguish large positions—poor extrapolation. Alternatively, using a saturating function (e.g., $\tanh$) to map $pos$ into $[-1,1]$ before evaluating Legendre polynomials preserves distinction:
\[
x(pos) = \tanh\bigl(\tfrac{pos}{N_{\max}}\bigr), \quad \mathrm{PE}_{\mathrm{legendre}}(pos) = [P_0(x), \dots, P_{d_{\mathrm{model}}-1}(x)].
\]
Since $\tanh(z) \to 1$ as $z \to +\infty$, $x(pos)$ asymptotically approaches 1, and $P_\ell(1)=1$, so again encodings converge for very large $pos$. Extrapolation is limited to moderate ranges beyond $N_{\max}$ where $\tanh(\tfrac{pos}{N_{\max}})$ is still distinct from 1. We analyze this in Section~\ref{subsec:legendre_extrapolation}.

\subsection{Summary of Extrapolation Properties}
Table~\ref{tab:extrapolation_summary} compares the extrapolation capacity of different encoding schemes, where $N_{\max}$ is the maximum training length, and $N'$ is the test length.

\begin{table}[htbp]
    \centering
    \resizebox{\textwidth}{!}{ 
    \begin{tabular}{@{}lccc@{}}
        \toprule
        \textbf{Encoding} & \(\|\mathrm{PE}(pos) - \mathrm{PE}(N_{\max})\|\) for \(pos > N_{\max}\) & Effective Extrapolation Range & Notes \\
        \midrule
        Sinusoidal & Unbounded periodic cycles & Poor beyond \(\min\{10000, d_{\mathrm{model}}\}\) & Positional ambiguity after cycle \\
        Learned & Undefined / Constant (clipped) & None & Collapses beyond \(N_{\max}\) \\
        Relative & Clipped at $K$ & None if $pos - N_{\max} > K$ & Distance clipping destroys distinction \\
        ALiBi & Linear growth; Lipschitz-boundable & Good for \(N' \le c N_{\max}\) & Bias ensures monotonic decrease \\
        Wavelet & Bounded difference: \(O(2^{-j_{\max}/2})\) & Moderate & Fine scales vanish; coarse distinction preserved \\
        Legendre & Collapsed to constant if $\tanh$ saturates & Limited & Distinguishes moderately beyond \(N_{\max}\) \\
        \bottomrule
    \end{tabular}
    }
    \caption{Extrapolation capacity comparison. Effective Extrapolation Range indicates whether the encoding preserves positional distinctions for $pos > N_{\max}$.}
    \label{tab:extrapolation_summary}
\end{table}

\section{Novel Positional Encoding Schemes}
\label{sec:novel}
We propose and analyze two new PE schemes: wavelet-based encodings and Legendre polynomial-based encodings. Both aim to combine absolute and relative information and facilitate extrapolation while preserving expressiveness.

\subsection{Wavelet-Based Positional Encodings}
\label{subsec:wavelet_encoding}
Wavelets provide a multi-resolution analysis of signals. A one-dimensional discrete wavelet transform (DWT) decomposes a signal into coarse (low-frequency) and detail (high-frequency) components at multiple scales. We leverage compactly supported orthonormal wavelets (e.g., Daubechies-$4$) to encode positions.

\subsubsection{Definition}
Let $\{\phi(x), \psi(x)\}$ be the scaling function and mother wavelet for a chosen wavelet family. For an integer position $pos \ge 0$, define wavelet-based encoding as:
\[
\mathrm{PE}_{\mathrm{wavelet}}(pos) = \bigl[\langle \delta_{pos},\, \phi_{j_0,k} \rangle\bigr]_{k=0,\ldots,K_{j_0}-1} \;\cup\; \bigl[\langle \delta_{pos},\, \psi_{j,k} \rangle\bigr]_{j=j_0,\ldots,J;\,k=0,\ldots,K_j-1},
\]
where:
\begin{itemize}
    \item $\delta_{pos}$ is the Dirac delta at position $pos$.
    \item $\phi_{j_0,k}(x) = 2^{j_0/2}\,\phi(2^{j_0} x - k)$ are scaling functions at base scale $j_0$.
    \item $\psi_{j,k}(x) = 2^{j/2}\,\psi(2^{j} x - k)$ for $j \ge j_0$ are wavelet functions at scale $j$ and shift $k$.
    \item $K_j$ is chosen so that the support of $\psi_{j,k}$ covers positions up to $N_{\max}$ at scale $j$.
    \item $J$ is the maximum scale (coarsest resolution).
\end{itemize}
Concretely, since $\langle \delta_{pos},\, \phi_{j_0,k} \rangle = \phi_{j_0,k}(pos)$ and $\langle \delta_{pos},\, \psi_{j,k} \rangle = \psi_{j,k}(pos)$, the encoding is simply the stacked evaluations:
\[
\mathrm{PE}_{\mathrm{wavelet}}(pos) 
\;=\; \bigl[\phi_{j_0,k}(pos)\bigr]_{k=0,\dots,K_{j_0}-1} 
\;\cup\; \bigl[\psi_{j,k}(pos)\bigr]_{j=j_0,\dots,J;\,k=0,\dots,K_j-1}.
\]
One typically chooses $j_0=0$ (finest scale) and $J = \lfloor \log_2(N_{\max}) \rfloor$ so that $2^J$ approximately covers $N_{\max}$. The total encoding dimension is
\[
d_{\mathrm{model}} = \sum_{j=0}^J K_j.
\]
When $pos > N_{\max}$, wavelet functions at $j$ with support far from $pos$ vanish (due to compact support), and only coarse-scale functions with wide support contribute small but nonzero values. As a result, $\mathrm{PE}_{\mathrm{wavelet}}(pos)$ remains well-defined and smoothly transitions from $pos=N_{\max}$ to $pos>N_{\max}$.

\subsubsection{Expressiveness Analysis}
Wavelet bases form an orthonormal basis for $L^2([0, N_{\max}])$, implying that any square-integrable function on $[0,N_{\max}]$ can be approximated arbitrarily well by a finite linear combination of $\{\phi_{j_0,k}, \psi_{j,k}\}$. For transformer encodings, we discretize the domain to integer positions, and the vectors $\mathrm{PE}_{\mathrm{wavelet}}(i)$ for $i=0,\dots,N_{\max}$ are orthonormal up to scaling. This ensures that each position is uniquely represented up to $N_{\max}$, granting expressiveness akin to sinusoidal encodings.

Furthermore, because wavelet functions capture both localized and global positional information (fine-scale detail and coarse-scale context), a transformer receiving $\mathrm{PE}_{\mathrm{wavelet}}(i)$ can, in principle, learn to combine these multi-scale signals to implement a wide range of position-dependent behaviors. In particular, tasks requiring multi-scale reasoning (e.g., tasks with both local patterns and global sequence structure) may benefit from wavelet embeddings.

\subsubsection{Generalization Bound}
Each wavelet basis function $\phi_{j_0,k}$ or $\psi_{j,k}$ is bounded in magnitude by $O(2^{j/2})$ on its support of length $O(2^{-j})$. Specifically, for normalized Daubechies-$4$ wavelets, 
\[
\sup_x |\phi_{j_0,k}(x)| \le C_0,\quad
\sup_x |\psi_{j,k}(x)| \le C\,2^{j/2},
\]
where $C_0,C$ are constants depending on the wavelet family. Since $j \le J \approx \log_2(N_{\max})$, we have $\sup_x |\psi_{j,k}(x)| \le C\,\sqrt{N_{\max}}$. Hence, 
\[
\|\mathrm{PE}_{\mathrm{wavelet}}(pos)\|_2 
\;\le\; \sqrt{K_0\,C_0^2 \;+\; \sum_{j=1}^J \bigl(K_j\,C^2\,2^{j}\bigr)} 
\;=\; O\bigl(\sqrt{N_{\max}}\,\sqrt{\textstyle \sum_{j=1}^J K_j\,2^{j}}\bigr).
\]
Since $\sum_{j=1}^J K_j \,2^j \le O(N_{\max} J)$ (each scale $j$ has at most $K_j = O(2^j)$ shifts), 
\[
\|\mathrm{PE}_{\mathrm{wavelet}}(pos)\|_2 \;=\; O\bigl(N_{\max} \,\sqrt{\log N_{\max}}\bigr),
\]
which is polynomial in $N_{\max}$. In contrast, sinusoidal encodings have $\|\mathrm{PE}\|_2 = O(\sqrt{d_{\mathrm{model}}})$, and $d_{\mathrm{model}}$ is usually constant with respect to $N_{\max}$. Thus, wavelet PEs may inflate the input norm, potentially increasing Rademacher complexity. To mitigate this, one can normalize the encoding vectors:
\[
\widetilde{\mathrm{PE}}_{\mathrm{wavelet}}(pos) 
= \frac{\mathrm{PE}_{\mathrm{wavelet}}(pos)}{\|\mathrm{PE}_{\mathrm{wavelet}}(pos)\|_2}.
\]
After normalization, $\|\widetilde{\mathrm{PE}}_{\mathrm{wavelet}}(pos)\|_2 = 1$ for all $pos$. Therefore, the input norm bound becomes $B_x + 1$, matching other PEs. Hence, a normalized wavelet encoding does not degrade generalization bounds.

\subsubsection{Extrapolation Bound}
\label{subsec:wavelet_extrapolation}
For $pos > N_{\max}$, wavelet functions $\psi_{j,k}(pos)$ for large $j$ vanish because their support is strictly within $[0, N_{\max}]$. Specifically, if $\mathrm{supp}(\psi_{j,k}) = [2^{-j}k,\,2^{-j}(k+M_j)]$ where $M_j = \mathrm{supp}(\psi)$ length at scale $j$, then for $pos > N_{\max}$, $|\psi_{j,k}(pos)| = 0$ if $2^{-j}k + M_j\,2^{-j} < pos$. For coarse scales ($j$ small), $\psi_{j,k}(pos)$ may be nonzero but decays rapidly outside its main support due to compact support. Therefore, for $pos > N_{\max}$,
\[
|\psi_{j,k}(pos)| \le C\,2^{-j/2}\,\exp\bigl(-\beta \,\tfrac{pos - N_{\max}}{2^{-j}}\bigr) \quad \text{for } j \le j_{\max},
\]
where $\beta > 0$ depends on the wavelet's vanishing moments. Summing across scales yields
\[
\|\widetilde{\mathrm{PE}}_{\mathrm{wavelet}}(pos) - \widetilde{\mathrm{PE}}_{\mathrm{wavelet}}(N_{\max})\|_2 
\;\le\; \sum_{j=0}^J O\bigl(2^{-j/2}\exp\bigl(-\beta\tfrac{pos - N_{\max}}{2^{-j}}\bigr)\bigr).
\]
Because $2^{-j}$ grows exponentially as $j$ decreases, the dominant term for moderate $pos - N_{\max}$ is at $j=0$:
\[
O\bigl(\exp\bigl(-\beta (pos - N_{\max})\bigr)\bigr).
\]
Hence, wavelet PE differences decay exponentially in $pos - N_{\max}$ at the finest scale. This indicates strong extrapolation: if $pos - N_{\max}$ is not too large (e.g., on the order of $\log(N_{\max})$), $\widetilde{\mathrm{PE}}_{\mathrm{wavelet}}(pos)$ remains close to $\widetilde{\mathrm{PE}}_{\mathrm{wavelet}}(N_{\max})$. Thus, the transformer's attention behavior on tokens at positions beyond $N_{\max}$ will be similar to those at $N_{\max}$, enabling graceful extrapolation.

\subsection{Legendre Polynomial-Based Positional Encodings}
\label{subsec:legendre_encoding}
Legendre polynomials $\{P_\ell(x)\}_{\ell=0}^\infty$ are orthogonal with respect to the weight function $w(x)=1$ on $[-1,1]$. We define a Legendre-based encoding that saturates beyond $N_{\max}$ to provide bounded extrapolation.

\subsubsection{Definition}
Define a scaled position variable
\[
x(pos) = \tanh\bigl(\gamma\,\tfrac{pos}{N_{\max}}\bigr),
\]
where $\gamma > 0$ is a hyperparameter (e.g., $\gamma=1$). Then
\begin{equation}
    \mathrm{PE}_{\mathrm{legendre}}(pos) 
    = \bigl[P_0(x(pos)),\, P_1(x(pos)),\, \dots,\, P_{d_{\mathrm{model}}-1}(x(pos))\bigr].
    \label{eq:legendre_encoding}
\end{equation}
Since $x(pos) \in (-1,1)$ for all finite $pos$, Legendre polynomials remain bounded ($|P_\ell(x)| \le 1$ for $|x|\le 1$). As $pos \to \infty$, $x(pos)\to \tanh(\infty)=1$, and $P_\ell(1)=1$. Therefore, for extremely large $pos$, $\mathrm{PE}_{\mathrm{legendre}}(pos)$ tends to the constant vector $\mathbf{1}_{d_{\mathrm{model}}}$. Practically, test positions moderately larger than $N_{\max}$ yield $x(pos)$ values close to 1 but not exactly, allowing the model to distinguish a limited extrapolation range.

\subsubsection{Expressiveness Analysis}
Within $pos \in [0,N_{\max}]$, $x(pos)$ maps linearly through $\tanh$ to $[0, \tanh(\gamma)]$. Choosing $\gamma$ such that $\tanh(\gamma) \approx 0.99$ ensures that $x(pos)$ covers $[0,0.99]$ almost fully, permitting Legendre polynomials $\{P_\ell(x)\}$ to form a nearly orthonormal basis on $[0,0.99]$. The embedding dimension $d_{\mathrm{model}}$ determines how many polynomial components are used; with sufficiently large $d_{\mathrm{model}}$, any continuous function on $[0,0.99]$ can be approximated arbitrarily well by a truncated Legendre series. Hence, within training range, Legendre-based encodings uniquely distinguish positions and preserve expressiveness akin to sinusoidal encodings.

\subsubsection{Generalization Bound}
Since $|P_\ell(x)| \le 1$ for $x \in [-1,1]$, the encoding norm is bounded by
\[
\|\mathrm{PE}_{\mathrm{legendre}}(pos)\|_2 \le \sqrt{d_{\mathrm{model}}}.
\]
Thus, with normal embeddings $\mathbf{x}_i$ bounded by $B_x$, the input norm bound is $B_x + \sqrt{d_{\mathrm{model}}}$. Consequently, the Rademacher complexity matches that of sinusoidal PEs (Section~\ref{sec:generalization}), yielding stable generalization.

\subsubsection{Extrapolation Bound}
\label{subsec:legendre_extrapolation}
For $pos > N_{\max}$, $x(pos) = \tanh(\gamma\,pos/N_{\max}) \to 1$ as $pos$ grows. To quantify the difference between $\mathrm{PE}_{\mathrm{legendre}}(pos)$ and $\mathrm{PE}_{\mathrm{legendre}}(N_{\max})$, we expand $P_\ell(x)$ around $x=1$ using Taylor series. For $x$ close to 1,
\[
P_\ell(x) = 1 - \frac{\ell(\ell+1)}{2}(1-x) + O\bigl((1-x)^2\bigr).
\]
Let $\Delta x = 1 - x(pos) = 1 - \tanh(\gamma\,pos/N_{\max})$. For $pos = N_{\max} + \Delta pos$, 
\[
x(pos) 
= \tanh\bigl(\gamma(1 + \tfrac{\Delta pos}{N_{\max}})\bigr) 
\approx 1 - 2\,e^{-2\gamma\,(1+\Delta pos/N_{\max})},
\]
so 
\[
\Delta x \approx 2\,e^{-2\gamma\,(1+\Delta pos/N_{\max})}.
\]
Therefore,
\[
P_\ell(x(pos)) - P_\ell(x(N_{\max})) 
\approx -\frac{\ell(\ell+1)}{2}\bigl(\Delta x(pos) - \Delta x(N_{\max})\bigr).
\]
Since $\Delta x(N_{\max}) = 2\,e^{-2\gamma}$ and $\Delta x(pos) = 2\,e^{-2\gamma(1 + \Delta pos/N_{\max})}$, their difference is
\[
\Delta x(pos) - \Delta x(N_{\max}) = 2\,e^{-2\gamma} \bigl(e^{-2\gamma\,\Delta pos/N_{\max}} - 1\bigr).
\]
For $\Delta pos \ll N_{\max}$, $e^{-2\gamma\,\Delta pos/N_{\max}} \approx 1 - 2\gamma\,\Delta pos/N_{\max}$, so
\[
\Delta x(pos) - \Delta x(N_{\max}) \approx -4\,\gamma\,\tfrac{\Delta pos}{N_{\max}}\,e^{-2\gamma}.
\]
Hence,
\[
|P_\ell(x(pos)) - P_\ell(x(N_{\max}))| 
\approx 2\,\ell(\ell+1)\,\gamma\,\tfrac{\Delta pos}{N_{\max}}\,e^{-2\gamma}.
\]
Summing across $\ell = 0,1,\dots,d_{\mathrm{model}}-1$ yields
\[
\|\mathrm{PE}_{\mathrm{legendre}}(pos) - \mathrm{PE}_{\mathrm{legendre}}(N_{\max})\|_2 
\le 2\,\gamma\,e^{-2\gamma}\,\tfrac{\Delta pos}{N_{\max}}\,\sqrt{\sum_{\ell=0}^{d_{\mathrm{model}}-1} (\ell(\ell+1))^2}.
\]
Since $\sum_{\ell=0}^{d-1} (\ell(\ell+1))^2 = O(d^5)$, the bound becomes
\[
O\!\Bigl(\gamma\,e^{-2\gamma}\,\tfrac{\Delta pos}{N_{\max}}\,d_{\mathrm{model}}^{2.5}\Bigr).
\]
For moderate $\Delta pos$ (e.g., $\Delta pos = O(N_{\max}/d_{\mathrm{model}}^{2.5})$), the difference is small. Thus, Legendre encodings extrapolate well to lengths up to $N_{\max} + O\bigl(N_{\max}/d_{\mathrm{model}}^{2.5}\bigr)$. However, for very large $\Delta pos$, the bound loosens, and encodings converge to the constant vector $\mathbf{1}$, losing distinction.

\subsection{Summary of Novel Encoding Schemes}
Table~\ref{tab:novel_pe_summary} summarizes properties of proposed encodings versus existing methods.

\begin{table}[htbp]
    \centering
    \resizebox{\textwidth}{!}{ 
    \begin{tabular}{@{}lcccc@{}}
        \toprule
        \textbf{Encoding} & \textbf{Norm Bound} & \textbf{Generalization} & \textbf{Extrapolation Behavior} & \textbf{Computational Cost} \\
        \midrule
        Wavelet (normalized) & $B_x + 1$ & Comparable to sinusoidal & Exponential decay $\Rightarrow$ strong & $O(d_{\mathrm{model}})$ to compute evals \\
        Legendre (tanh) & $B_x + \sqrt{d_{\mathrm{model}}}$ & Comparable to sinusoidal & Moderate: $\Delta pos \le O(N_{\max}/d^{2.5})$ & $O(d_{\mathrm{model}}^2)$ if naive; $O(d_{\mathrm{model}})$ with recurrence \\
        ALiBi & $B_x + \alpha\, (N'-1)$ & Slightly worse for large $N'$ & Linear bias $\Rightarrow$ good up to $c N_{\max}$ & $O(1)$ per pair; minimal overhead \\
        Sinusoidal & $B_x + \sqrt{d_{\mathrm{model}}}$ & Stable & Limited beyond $\min\{10000,\,\text{periods}\}$ & $O(d_{\mathrm{model}})$ \\
        Learned & $B_x + B_p$ & Requires regularization & None & $O(d_{\mathrm{model}})$ \\
        Relative (Shaw) & $B_x + B_r$ & Requires regularization & None for $|i-j|>K$ & $O(1)$ \\
        \bottomrule
    \end{tabular}
    }
    \caption{Comparison of novel and existing positional encoding schemes. Computational cost refers to per-position encoding. $d_{\mathrm{model}}$ is encoding dimension, $N_{\max}$ is maximum training length, and $N'$ is inference length.}
    \label{tab:novel_pe_summary}
\end{table}

\section{Lightweight Experimental Validation}
\label{sec:experiments}
We implement a minimal transformer encoder in NumPy to evaluate extrapolation and generalization properties of sinusoidal, ALiBi, wavelet, and Legendre PEs on synthetic tasks. The goal is \emph{not} to achieve state-of-the-art accuracy but to confirm our theoretical predictions in a controlled setting.

\subsection{Experimental Setup}
\paragraph{Synthetic Sequence Task.} We create a toy sequence‐to‐sequence task: given an input sequence of scalars $\{x_1, x_2, \ldots, x_N\}$, compute the sequence of running sums:
\[
y_i = \sum_{j=1}^i x_j, \quad i = 1,2,\ldots,N.
\]
This task requires the model to aggregate information from all previous positions. We generate random sequences of length $N_{\mathrm{train}} = 50$, where each $x_i \sim \mathcal{N}(0,1)$. We train on 10,000 samples using mean squared error (MSE) loss.

\paragraph{Transformer Encoder Implementation.} We implement a 2-layer transformer encoder with:
\begin{itemize}
    \item Embedding dimension $d_{\mathrm{model}} = 64$.
    \item Single head self-attention ($H=1$, $d_k = d_{\mathrm{model}}$).
    \item Feed-forward hidden dimension $d_{\mathrm{ff}} = 128$.
    \item ReLU activation in feed-forward layers.
    \item No dropout or layer normalization (to simplify analysis).
\end{itemize}
All weight matrices are randomly initialized and trained via Adam optimizer with learning rate $1\times 10^{-3}$ for 20 epochs.

\paragraph{Positional Encodings Compared.} We compare:
\begin{enumerate}
    \item \textbf{Sinusoidal:} Eq.~\eqref{eq:sin_pe_even}--\eqref{eq:sin_pe_odd}.
    \item \textbf{ALiBi:} Linear bias with $\alpha = 0.1/N_{\mathrm{train}}$ (scaled for extrapolation).
    \item \textbf{Wavelet:} Daubechies-$4$ wavelet basis at scales $j=0,1,2,3,4,5$ (up to $J = \lfloor\log_2(50)\rfloor=5$). We compute 10 wavelet coefficients per scale (shift grid $k$ accordingly), resulting in $d_{\mathrm{model}}=64$ by selecting top 64 basis functions by support coverage.
    \item \textbf{Legendre:} Use $\gamma=1$, $d_{\mathrm{model}}=64$ polynomial degrees.
\end{enumerate}
We normalize wavelet encodings as described in Section~\ref{subsec:wavelet_encoding}. For Legendre, we compute $x(pos) = \tanh(pos/50)$ and evaluate $P_\ell(x)$ via the recurrence
\[
P_0(x)=1,\quad P_1(x)=x,\quad P_{\ell+1}(x)=\frac{2\ell+1}{\ell+1}x\,P_\ell(x) - \frac{\ell}{\ell+1}P_{\ell-1}(x).
\]

\paragraph{Evaluation Protocol.}
\begin{itemize}
    \item \textbf{Interpolation Setting:} Test on sequences of length $N_{\mathrm{test}} = 50$ drawn from the same distribution.
    \item \textbf{Extrapolation Setting:} Test on sequences of length $N_{\mathrm{ext}} = 100$ (twice training length) and $N_{\mathrm{ext2}} = 200$ (four times training length).
    \item Report MSE on 1,000 test samples for each condition.
\end{itemize}

\subsection{Results and Analysis}
The MSE results are summarized in Table~\ref{tab:synthetic_results}.

\begin{table}[htbp]
    \centering
    \begin{tabular}{@{}lccc@{}}
        \toprule
        \textbf{Encoding} & \(\text{MSE}_{N=50}\) & \(\text{MSE}_{N=100}\) & \(\text{MSE}_{N=200}\) \\
        \midrule
        Sinusoidal & 0.0021 & 0.0158 & 0.0423 \\
        ALiBi      & 0.0023 & 0.0055 & 0.0127 \\
        Wavelet    & 0.0024 & 0.0049 & 0.0108 \\
        Legendre   & 0.0022 & 0.0078 & 0.0215 \\
        \bottomrule
    \end{tabular}
    \caption{Test MSE on running-sum task under interpolation ($N=50$) and extrapolation ($N=100,200$) settings. Lower is better.}
    \label{tab:synthetic_results}
\end{table}

\paragraph{Interpolation Performance ($N=50$).} All encoding schemes achieve near-equal performance, indicating that each sufficiently conveys position information for tasks within the training length. Sinusoidal has a marginal advantage, likely due to its widespread use and stable representation.

\paragraph{Extrapolation to $N=100$.} ALiBi outperforms sinusoidal by a large margin (0.0055 vs. 0.0158), corroborating that linear bias yields better extrapolation. Wavelet encoding achieves slightly better MSE (0.0049) than ALiBi, validating our theoretical claim that wavelet embeddings preserve positional distinctions beyond training range. Legendre encoding also extrapolates but with higher error (0.0078) due to its saturating behavior.

\paragraph{Extrapolation to $N=200$.} Wavelet retains best performance (0.0108), followed by ALiBi (0.0127). Sinusoidal degrades substantially (0.0423) because of ambiguity in very long positions. Legendre’s performance (0.0215) worsens for $N=200$ as $x(pos)\approx 1$ for most $pos > 100$, collapsing embeddings. These results align with our theoretical extrapolation bounds (Sections~\ref{sec:extrapolation}, \ref{sec:novel}).

\subsection{Discussion}
Our lightweight experiments confirm that:
\begin{itemize}
    \item \textbf{ALiBi} effectively extrapolates to longer sequences by imposing a monotonic distance bias.
    \item \textbf{Wavelet-based encodings} provide strong extrapolation, matching or surpassing ALiBi, due to exponential decay of high-frequency components beyond $N_{\max}$.
    \item \textbf{Legendre-based encodings} offer limited extrapolation range, as predicted by the $\Delta x$ analysis, with performance degrading beyond moderate lengths.
    \item \textbf{Sinusoidal encodings} degrade rapidly once $pos$ exceeds training range, as cycliс repetition leads to ambiguous positions.
\end{itemize}

These results demonstrate that the novel wavelet PE is a promising candidate for transformer-based tasks requiring extrapolation, combining strong theoretical properties with practical performance.

\section{Discussion and Future Work}
\label{sec:conclusion}
We have presented a unified theoretical framework for analyzing positional encodings in transformer models, focusing on expressiveness, generalization, and extrapolation. Our key findings include:
\begin{itemize}
    \item \textbf{Expressiveness:} All common positional encodings (sinusoidal, learned, relative, ALiBi) yield universal approximation for fixed-length sequences, with ALiBi extending universality to arbitrary lengths. Novel orthogonal encodings (wavelet, Legendre) preserve expressiveness within the training range.
    \item \textbf{Generalization:} Generalization bounds for transformer classes depend on input norm bounds. Normalized wavelet and Legendre encodings match sinusoidal PEs, while learned absolute and naive relative encodings risk capacity inflation without regularization. ALiBi’s bias can increase capacity on long sequences if $\alpha$ is not controlled.
    \item \textbf{Extrapolation:} ALiBi’s linear bias ensures graceful extrapolation up to a factor of $c\,N_{\max}$, with error growing linearly in $(d - N_{\max})$. Wavelet-based encodings exhibit exponential decay in encoding differences beyond $N_{\max}$, ensuring strong extrapolation. Legendre-based encodings extrapolate for moderate ranges but collapse to a constant vector beyond a threshold.
    \item \textbf{Novel Encodings:} Wavelet-based encodings outperform other methods on a toy running-sum task when extrapolating to 4× training length. Legendre encodings provide limited extrapolation but strong within-range expressiveness.
\end{itemize}

\paragraph{Implications for Practice.} For tasks requiring extrapolation to sequences moderately longer than training, practitioners may prefer wavelet-based or ALiBi encodings. Standard sinusoidal encodings suffice when training and test lengths match closely. Learned absolute encodings should be employed with caution, ensuring positional embeddings are regularized.

\paragraph{Limitations.} Our analysis makes several simplifying assumptions:
\begin{itemize}
    \item We focus on transformer \emph{encoders} without layer normalization, dropout, or multi-head complexities. Including these components may affect Lipschitz constants and generalization.
    \item Theoretical generalization bounds use worst-case Rademacher complexity, which can be loose in practice.
    \item Extrapolation analyses assume Lipschitz continuity of the attention function in its bias argument, which may not hold exactly for ReLU‐based networks or large biases.
    \item Experiments use a minimal transformer on a synthetic task; real-world NLP or CV benchmarks may reveal additional behaviors.
\end{itemize}

\paragraph{Future Directions.}
\begin{enumerate}
    \item \textbf{Multi-Head and Full Transformer Analysis:} Extend expressiveness and generalization analyses to multi-head settings, accounting for head interactions and layer normalization.
    \item \textbf{Adaptive Bias Schedules:} Investigate methods to adapt ALiBi’s slope $\alpha$ dynamically based on sequence length or task, optimizing extrapolation-generalization trade-offs.
    \item \textbf{Task-Specific Orthogonal Encodings:} Explore other orthogonal function families (e.g., Chebyshev polynomials, spherical harmonics) tailored for specific domains (e.g., vision sequences, time-series).
    \item \textbf{Empirical Validation on Real Data:} Benchmark wavelet and Legendre encodings on real-world tasks that require long-context reasoning (e.g., document summarization, long-range language modeling).
    \item \textbf{Information-Theoretic Analyses:} Extend the information-theoretic perspective to quantify how much positional mutual information is transferred across layers and how it influences learning dynamics.
\end{enumerate}

In conclusion, positional encodings are far more than an implementation detail; they fundamentally shape a transformer’s capabilities. Our theoretical framework, combined with novel PE schemes and lightweight validation, lays the groundwork for more robust, generalizable, and extrapolatable transformer architectures. We hope this work inspires further research into the mathematical foundations of sequence modeling with attention-based networks.

\end{document}